\pdfoutput=1

\documentclass[11pt]{article}

\usepackage[]{acl}

\usepackage{times}
\usepackage{latexsym}

\usepackage[T1]{fontenc}

\usepackage[utf8]{inputenc}

\usepackage{microtype}

\usepackage{multirow,multicol}
\usepackage{booktabs}
\usepackage{amsmath}
\usepackage{amssymb}
\usepackage{adjustbox}
\usepackage{semtrans}
\usepackage{svg}
\usepackage{pdfpages}
\usepackage{graphicx}
\usepackage{caption}
\usepackage{subcaption}
\usepackage{tipa}

\title{{A}mericas{NLI}: {E}valuating {Z}ero-shot {N}atural {L}anguage {U}nderstanding\\of {P}retrained {M}ultilingual {M}odels in {T}ruly {L}ow-resource {L}anguages}

\author{Abteen Ebrahimi${ }^{\diamondsuit}$ \quad
Manuel Mager${ }^{\spadesuit}$ \quad
Arturo Oncevay${ }^{\heartsuit}$ \quad
Vishrav Chaudhary${ }^{\S}$  \\
\textbf{Luis Chiruzzo${ }^{\triangle}$ \enskip
Angela Fan${ }^{\nabla}$ \enskip
John E. Ortega${ }^{\Omega}$ \enskip
Ricardo Ramos${ }^{\eta}$ \enskip
Annette Rios${ }^{\psi}$ } \\
\textbf{Ivan Meza-Ruiz${ }^{\sharp}$ \enskip
Gustavo A. Giménez-Lugo${ }^{\clubsuit}$ \enskip
Elisabeth Mager${ }^{\sharp}$ \enskip
Graham Neubig${ }^{\bowtie}$ } \\
\textbf{Alexis Palmer${ }^{\diamondsuit}$  \enskip
Rolando Coto-Solano${ }^{\mho}$ \enskip
Ngoc Thang Vu${ }^{\spadesuit}$  \enskip
Katharina Kann${ }^{\diamondsuit}$} \\
${ }^{\bowtie}$Carnegie Mellon University \quad
${ }^{\mho}$Dartmouth College \quad
${ }^{\S}$Microsoft Turing \\
${ }^{\nabla}$Facebook AI Research \enskip
${}^{\Omega}$New York University \enskip
${ }^{\triangle}$Universidad de la República, Uruguay \\
${}^{\eta}$Universidad Tecnológica de Tlaxcala \quad
${ }^{\sharp}$Universidad Nacional Autónoma de México \\
${ }^{\clubsuit}$Universidade Tecnológica Federal do Paraná \quad
${ }^{\diamondsuit}$University of Colorado Boulder \\
${ }^{\heartsuit}$University of Edinburgh \quad
${ }^{\spadesuit}$University of Stuttgart \quad
${ }^{\psi}$University of Zurich \quad}

\begin{document}
\maketitle
\begin{abstract} 
Pretrained multilingual models are able to perform cross-lingual transfer in a zero-shot setting, even for languages unseen during pretraining. However, prior work evaluating performance on unseen languages has largely been limited to low-level, syntactic tasks, and it remains unclear if zero-shot learning of high-level, semantic tasks is possible for unseen languages. To explore this question, we present AmericasNLI, an extension of XNLI \cite{conneau2018xnli} to 10 Indigenous languages of the Americas.
We conduct experiments with XLM-R, testing multiple zero-shot and translation-based approaches.
Additionally, we explore model adaptation via continued pretraining and provide an analysis of the dataset by considering hypothesis-only models. We find that XLM-R's zero-shot performance is poor for all 10 languages, with an average performance of 38.48\%. Continued pretraining offers improvements, with an average accuracy of 43.85\%. Surprisingly, training on poorly translated data by far outperforms all other methods with an accuracy of 49.12\%.
\end{abstract}

\section{Introduction}

Pretrained multilingual models such as XLM \cite{Lample2019CrosslingualLM}, multilingual BERT
\citep[mBERT;][]{Devlin2019}, and XLM-R \cite{Conneau2020UnsupervisedCR} achieve strong cross-lingual transfer results for many languages and natural language processing (NLP) tasks. However, there exists a discrepancy in terms of zero-shot performance between languages present in the pretraining data and those that are not: performance is generally highest for well-represented languages and decreases with less representation. Yet, even for unseen languages, performance is generally above chance, and model adaptation approaches have been shown to yield further improvements \cite{Muller2020WhenBU, Pfeiffer2020MADXAA,pfeiffer2020unks,wang-etal-2020-extending}.

Importantly, however, there are currently no datasets for high-level, semantic tasks which focus solely on low-resource languages. As these languages are most likely to be unseen to commonly used pretrained models, practically all work evaluating unseen language performance and language adaptation methods has been limited to low-level, syntactic tasks such as part-of-speech tagging, dependency parsing, and named-entity recognition \cite{Muller2020WhenBU, wang-etal-2020-extending}. This largely limits our ability to draw more general conclusions with regards to the zero-shot learning abilities of pretrained multilingual models for unseen languages.

\begin{table}[]
    \centering
    \small

    \begin{tabular}{lllrr}
        \toprule
        \textbf{Language} & \textbf{ISO} & \textbf{Family} & \textbf{Dev} & \textbf{Test} \\ \midrule
        Aymara & aym & Aymaran  & 743 & 750  \\
        Asháninka & cni & Arawak & 658 & 750  \\
        Bribri & bzd & Chibchan & 743 & 750   \\
        Guaraní & gn & Tupi-Guaraní & 743 & 750   \\
        Nahuatl & nah & Uto-Aztecan & 376 & 738   \\
        Otomí & oto & Oto-Manguean  & 222 & 748   \\
        Quechua & quy & Quechuan & 743 & 750   \\
        Rarámuri & tar & Uto-Aztecan & 743 & 750  \\
        Shipibo-Konibo & shp & Panoan & 743 &  750   \\
        Wixarika & hch & Uto-Aztecan & 743 & 750   \\

    \bottomrule
    \end{tabular}

    \caption{The languages in AmericasNLI, along with their ISO codes, language families, and dataset sizes.}
    \label{tab:languages}
\end{table}
In this work, we introduce AmericasNLI, an extension of XNLI \cite{conneau2018xnli} -- a natural language inference (NLI; cf. \S\ref{subsec:nli}) dataset covering 15 high-resource languages -- to 10 Indigenous languages spoken in the Americas: Ash\'{a}ninka, Aymara, Bribri, Guaraní, Nahuatl, Otomí, Quechua, Rar\'{a}muri, Shipibo-Konibo, and Wixarika. All of them are \emph{truly} low-resource languages: they have little to no digitally available labeled or unlabeled data, and they are not typically studied by the mainstream NLP community. The goal of this work is two-fold: First, we hope to increase the visibility of these languages by providing a portion of the resources necessary for NLP research. Second, we aim to allow for a more comprehensive study of multilingual model performance on unseen languages, where improvements will help extend the reach of NLP techniques to a larger set of languages. We are specifically interested in the following research questions: (1) Do pretrained multilingual models still perform above random chance for a high-level, semantic task in an unseen language? (2) Do methods aimed at adapting models to unseen languages -- previously exclusively evaluated on low-level, syntactic tasks -- also increase performance on NLI? (3) Are translation-based approaches effective for truly low-resource languages, where translation quality is typically very poor?\footnote{We provide a sample of sentence pairs in Table \ref{tab:translate_train_sample}.}

We experiment with XLM-R, both with and without model adaptation via continued pretraining on monolingual corpora in the target language. Our results show that the performance of XLM-R out-of-the-box is moderately above chance, and model adaptation leads to improvements of up to 5.86 percentage points. Training on machine-translated training data, however, results in an even larger performance gain of 11.13 percentage points over the corresponding XLM-R model without adaptation.
We further perform an analysis via experiments with hypothesis-only models, to examine potential artifacts which may have been inherited from XNLI and find that performance is above chance for most models, but still below that for using the full example.

AmericasNLI is publicly available\footnote{\url{https://github.com/abteen/americasnli}} and we hope that it will serve as a benchmark for measuring the zero-shot natural language understanding abilities of multilingual models for unseen languages. Additionally, we hope that our dataset will motivate the development of novel pretraining and model adaptation techniques which are suitable for truly low-resource
languages.

\section{Background and Related Work}
\subsection{Pretrained Multilingual Models}

Prior to the widespread use of pretrained transformer models, cross-lingual transfer was mainly achieved through word embeddings \cite{mikolov13,pennington-etal-2014-glove, bojanowski-etal-2017-enriching}, either by aligning monolingual embeddings into the same embedding space \cite{conneau2017word, lample2017unsupervised, grave2018learning} or by training multilingual embeddings \cite{Ammar2016MassivelyMW, Artetxe2019MassivelyMS}. Pretrained multilingual models represent the extension of multilingual embeddings to pretrained transformer models.

These models follow the standard pretraining--finetuning paradigm: they are first trained on unlabeled monolingual corpora from various languages (the \emph{pretraining languages}) and later finetuned on target-task data in a -- usually high-resource -- source language. Having been exposed to a variety of languages through this training setup, cross-lingual transfer results for these models are competitive with the state of the art for many languages and tasks. Commonly used models are mBERT \cite{Devlin2019}, which is pretrained on the Wikipedias of 104 languages with masked language modeling (MLM) and next sentence prediction (NSP), and XLM, which is trained on 15 languages and introduces the translation language modeling objective, which is based on MLM, but uses pairs of parallel sentences. XLM-R has improved performance over XLM, and trains on data from 100 different languages with only the MLM objective. Common to all models is a large shared subword vocabulary created using either BPE \cite{sennrich-etal-2016-neural} or SentencePiece \cite{kudo-richardson-2018-sentencepiece} tokenization.

\subsection{Evaluating Pretrained Multilingual Models}
Just as in the monolingual setting, where benchmarks such as GLUE \cite{wang-etal-2018-glue} and SuperGLUE \cite{wang2019superglue} provide a look into the performance of models across various tasks, multilingual benchmarks \cite{hu2020xtreme, liang-etal-2020-xglue} cover a wide variety of tasks involving sentence structure, classification, retrieval, and question answering. 

Additional work has been done examining what mechanisms allow multilingual models to transfer across languages \cite{pires-etal-2019-multilingual, wu-dredze-2019-beto}. \citet{Wu2020AreAL} examine transfer performance dependent on a language's representation in the pretraining data. For languages with low representation, multiple methods have been proposed to improve performance, including extending the vocabulary, transliterating the target text, and continuing pretraining before finetuning \cite{lauscher-etal-2020-zero, mbert-parsing, Muller2020WhenBU, Pfeiffer2020MADXAA, pfeiffer2020unks, wang-etal-2020-extending}. In this work, we focus on continued pretraining to analyze the performance of model adaptation for a high-level, semantic task.

\subsection{Natural Language Inference}
\label{subsec:nli}

Given two sentences, the \textit{premise} and the \textit{hypothesis}, the task of NLI consists of determining whether the hypothesis logically entails, contradicts, or is neutral to the premise.
The most widely used datasets for NLI in English are SNLI \cite{snli:emnlp2015} and MNLI \cite{mnli}. XNLI \cite{conneau2018xnli} is the multilingual expansion of MNLI to 15 languages, providing manually translated evaluation sets and machine-translated training sets. While datasets for NLI or the similar task of recognizing textual entailment exist for other languages \cite{Bos2009TextualEA, alabbas-2013-dataset, eichler-etal-2014-analysis, Amirkhani2020FarsTailAP}, their lack of similarity prevents a generalized study of cross-lingual zero-shot performance. This is in contrast to XNLI, where examples for all 15 languages are parallel. To preserve this property of XNLI, when creating AmericasNLI, we choose to translate Spanish XNLI as opposed to creating examples directly in the target language.

However, NLI datasets are not without issue: \citet{Gururangan2018AnnotationAI} show that artifacts from the creation of MNLI allow for models to classify examples depending on only the hypothesis, showing that models may not be
reasoning as expected.
Motivated by this, we provide further analysis of AmericasNLI in Section \ref{analysis} by comparing the performance of hypothesis-only models to models trained on full examples.

\section{AmericasNLI}
\subsection{Data Collection Setup}

AmericasNLI is the translation of a subset of XNLI \cite{conneau2018xnli}. As translators between Spanish and the target languages are more frequently available than those for English, we translate from the Spanish version. Additionally, some translators reported that code-switching is often used to describe certain topics, and, while many words without an exact equivalence in the target language are worked in through translation or interpretation, others are kept in Spanish. To minimize the amount of Spanish vocabulary in the translated examples, we choose sentences from genres that we judged to be relatively easy to translate into the target languages: ``face-to-face,'' ``letters,'' and ``telephone.'' We choose up to 750 examples from each of the development and test set, with exact counts for each language in Table \ref{tab:languages}.

\subsection{Languages}
We now discuss the languages in AmericasNLI. For additional background on previous NLP research on Indigenous languages of the Americas, we refer the reader to \citet{mager-etal-2018-challenges}. A summary of this information can be found in Table \ref{tab:language_summary}.

\paragraph{Aymara} Aymara is a polysynthetic Amerindian language spoken in Bolivia, Chile, and Peru by over two million people \cite{homola_petr}.
Aymara follows an SOV word order and has multiple dialects, including Northern and Southern Aymara,
spoken on the southern Peruvian shore of Lake Titicaca
as well as around La Paz and, respectively, in the eastern half of the Iquique province in northern Chile, the Bolivian department of Oruro, in northern Potosi, and southwest Cochabamba.
AmericasNLI examples are translated into the Central Aymara variant, specifically Aymara La Paz.

\begin{table*}[]
    \centering
    \setlength{\tabcolsep}{1.5pt}
    \small

    \begin{tabular}{cll}
        \toprule
        \textbf{Language} & \textbf{Premise} & \textbf{Hypothesis} \\
        \midrule

        {en} & And he said, Mama, I'm home.
         & He told his mom he had gotten home. \\
        \midrule
        {es} & Y él dijo: Mamá, estoy en casa.
        & Le dijo a su madre que había llegado a casa. \\
        \midrule
        {aym} & Jupax sanwa: Mamita, utankastwa.
         & Utar purinxtwa sasaw mamaparux sanxa \\
        \midrule
        {bzd} & \underline{E}n\underline{a} ie' iche:  ãm\textipa{\`{ĩ}}, ye' tso' ù \underline{a}.
         & I  ãm\textipa{\`{ĩ}} \underline{a} iché irir tö ye' dém\underline{i}n\underline{e} ù \underline{a}. \\
        \midrule
        {cni} & Iriori ikantiro: Ina, nosaiki pankotsiki.
        & Ikantiro iriniro yaretaja pankotsiki. \\
        \midrule
        {gn}  & Ha ha'e he'i: Mama, aime ógape.
        & He'íkuri isýpe oĝuahêhague hógape. \\
        \midrule
        {hch} & metá mik+ petay+: ne mama kitá nepa yéka.
        & yu mama m+pa+ p+ra h+awe kai kename yu kitá he nuakai. \\
        \midrule
        {nah} & huan yehhua quiihtoh: Nonantzin, niyetoc nochan
         & quiilih inantzin niehcoquia \\
        \midrule
        {oto} & xi nydi biênâ: maMe dimi an ngû
         & bimâbi o ini maMe guê o ngû \\
        \midrule
        {quy} & Hinaptinmi pay nirqa: Mamay wasipim kachkani.
         & Wasinman chayasqanmanta mamanta willarqa. \\
        \midrule
        {shp} & Jara neskata iki: tita, xobonkoriki ea.
        & Jawen tita yoiaia iki moa xobon nokota. \\
        \midrule
        {tar} & A’lí je aníli échiko: ku bitichí ne atíki Nana
         & Iyéla ku ruyéli, mapu bitichí ku nawáli. \\
    \bottomrule
    \end{tabular}

    \caption{A parallel example in AmericasNLI with the \textit{entailment} label.}
    \label{tab:examples}
\end{table*}
\paragraph{Asháninka}

Asháninka is an Amazonian language from the Arawak family, spoken by 73,567 people\footnote{\url{https://bdpi.cultura.gob.pe/pueblos/ashaninka}} in Central and
Eastern Peru, in a geographical region located between the eastern foothills of the
Andes and the western fringe of the Amazon basin \cite{mihas_2017}. 
Asháninka is an agglutinating and polysynthetic language with a VSO word order.

\paragraph{Bribri}
Bribri is a Chibchan language spoken by 7,000 people in Southern Costa Rica \cite{inec2011}. It has three dialects, and while it is still spoken by children, it is currently a vulnerable language \cite{unescoatlas,carlosvitalidad}.
Bribri is a tonal language with SOV word order. 
There are several orthographies which use different diacritics for the same phenomena, however even for researchers who use the same orthography, the Unicode encoding of similar diacritics differs amongst authors. 
Furthermore, the dialects of Bribri differ in their exact vocabularies, 
and there are phonological processes, like the deletion of unstressed vowels, which also change the tokens found in texts. 
As Bribri has only been a written language for about 40 years, existing materials have a large degree of idiosyncratic variation. These variations are standardized in AmericasNLI, which is written in the Amubri variant.

\paragraph{Guaraní}
Guaraní is spoken by between 6 to 10 million people in South America and roughly 3 million people use it as their main language, including more than 10 native nations in Paraguay, Brazil, Argentina, and Bolivia, along with Paraguayan, Argentinian, and Brazilian peoples.
According to the Paraguayan Census, in 2002 there were around 1.35 million monolingual speakers, which has since increased to around 1.5 million people \cite{dossantos2017,melia1992}.\footnote{\url{https://www.ine.gov.py/news/25-de-agosto-dia-del-Idioma-Guarani.php}}
Although the use of Guaraní as spoken language is much older, the first written record
dates to 1591 (Catechism) followed by the first dictionary in 1639 and linguistic descriptions in 1640.
The official grammar of Guaraní was approved in 2018. Guaraní is an agglutinative language, with ample use of prefixes and suffixes. 
 \begin{table*}[ht]
    \centering
    \small
    \setlength{\tabcolsep}{9pt}

    \begin{tabular}{cc|cccccccccc}
        \toprule
        & &\textbf{aym} & \textbf{bzd} & \textbf{cni} & \textbf{gn} & \textbf{hch} & \textbf{nah} & \textbf{oto} & \textbf{quy} & \textbf{shp} & \textbf{tar} \\
        \midrule
        \multirow{2}{*}{ChrF} & es$\rightarrow$XX & 0.19 & 0.08 & 0.10 & 0.22 & 0.13 & 0.18 & 0.06 & 0.33 & 0.14 & 0.05 \\
        & XX$\rightarrow$es & 0.09 & 0.06 & 0.09 & 0.14 & 0.07 & 0.10 & 0.06 & 0.14 & 0.09 & 0.08 \\
        \midrule
        \multirow{2}{*}{BLEU} & es$\rightarrow$XX & 0.30 & 0.54 & 0.03 & 3.26 & 3.18 & 0.33 & 0.01 & 1.58 & 0.34 & 0.01 \\
        & XX$\rightarrow$es & 0.04 & 0.01 & 0.01 &0.18 & 0.01 & 0.02  & 0.02 & 0.05 & 0.01 & 0.01 \\

    \bottomrule
    \end{tabular}

    \caption{Translation performance for all target languages. \textit{es$\rightarrow$XX} represents translating into the target language, which is used for translate-train, and \textit{XX$\rightarrow$es} represents translating into Spanish, used for translate-test.}
    \label{tab:translation_scores}
\end{table*}

\paragraph{Nahuatl}
Nahuatl belongs to the Nahuan subdivision of the Uto-Aztecan language family.
There are 30 recognized variants of Nahuatl spoken by over 1.5 million speakers across 
Mexico, where Nahuatl is recognized as an official language \cite{SEGOB2020sicNahuatl}.
Nahuatl is polysynthetic and agglutinative, 
and many sentences have an SVO word order or, for contrast and focus, a VSO order, and for emphasis, an SOV order \cite{macswan1998argument}. The translations in AmericasNLI belong to the Central Nahuatl (Náhuatl de la Huasteca) dialect. As there is a lack of consensus regarding the orthographic standard, the orthography is normalized to a version similar to Classical Nahuatl.

\paragraph{Otomí}
Otomí belongs to the Oto-Pamean language family and has nine linguistic variants with different regional self-denominations. Otomí is a tonal language following an SVO order,
and there are around 307,928 speakers spread across 7 Mexican states. In the state of Tlaxcala, the \textit{yuhmu} or \textit{ñuhmu} variant is spoken by fewer than 100 speakers, and we use this variant for the Otomí examples in AmericasNLI.

\paragraph{Quechua}
Quechua, or \textit{Runasimi}, is an Indigenous language family spoken 
primarily in the Peruvian Andes.
It is the most widely spoken pre-Columbian language family of the Americas, with around 8-10 million speakers. Approximately 25\% (7.7 million) of Peruvians speak a Quechuan language, and it is the co-official language in many regions of Peru.
There are multiple subdivisions of Quechua
, and AmericasNLI examples are translated into the standard version of Southern Quechua, Quechua Chanka, also known as Quechua Ayacucho, which is spoken in different regions of Peru and can be understood in different areas of other countries, such as Bolivia or Argentina. 
In AmericasNLI, the apostrophe and pentavocalism from other regions are not used.

\paragraph{Rarámuri}

Rarámuri, also known as \textit{Tarahumara}, which means \textit{light foot} \cite{INALI2017etnografia},
belongs to the Taracahitan subgroup of the Uto-Aztecan language family \cite{goddard1996introduction}, and is polysynthetic and agglutinative.
Rarámuri is an official language of Mexico, spoken mainly in the Sierra Madre Occidental region 
by a total of 89,503 speakers \cite{SEGOB2020sicRaramuri}. 
AmericasNLI examples are translated into the Highlands variant \cite{INALI2009catalogo}, and translation orthography and word boundaries are similar to \citet{caballero2008choguita}.

\paragraph{Shipibo-Konibo}
Shipibo-Konibo
is a Panoan language spoken by around 35,000 native speakers in the Amazon region of Peru. 
Shipibo-Konibo uses an SOV word order \cite{Faust-1973} and postpositions \cite{vasquez-etal-2018-toward}.
The translations in AmericasNLI make use of the official alphabet and standard writing supported by the Ministry of Education in Peru.

\paragraph{Wixarika}
The Wixarika, or \textit{Huichol}, language, meaning \textit{the language of the doctors and healers} \cite{lumholtz2011unknown}, 
is a language in the Corachol subgroup of the Uto-Aztecan language family \cite{campbell2000american}. Wixarika is a national language of Mexico with four variants
, spoken by a total of around 47,625 speakers \cite{SEGOB2020sicWixarika}. Wixarika is a polysynthetic language 
and follows an SOV word order. 
Translations in AmericasNLI are in Northern Wixarika and use an orthography common among native speakers \cite{mager2017traductor}.

\section{Experiments}
In this section, we detail the experimental setup we use to evaluate the performance of various approaches on
AmericasNLI.

\subsection{Zero-Shot Learning}
\begin{table*}[]
    \centering
    \small
    \setlength{\tabcolsep}{2pt}

    \begin{adjustbox}{width=\textwidth}
    \renewcommand{\arraystretch}{1.3}
    \begin{tabular}{l|cccccccccc|c}
        \toprule
           & \textbf{aym} & \textbf{bzd} & \textbf{cni} & \textbf{gn} & \textbf{hch} & \textbf{nah} & \textbf{oto} & \textbf{quy} & \textbf{shp} & \textbf{tar} & \textbf{Avg.}\\
        \midrule
        Majority baseline & 33.33 & 33.33 & 33.33 & 33.33 & 33.33 & 33.47 & 33.42 & 33.33 & 33.33 & 33.33 & -  \\
        \midrule
        \multicolumn{1}{l}{\textit{Zero-shot}} \\
        \midrule
         XLM-R (en) & 36.13\tiny{$\pm$0.88} & 39.65\tiny{$\pm$0.89} & 37.91\tiny{$\pm$0.82} & 39.47\tiny{$\pm$1.14} & 37.20\tiny{$\pm$1.32} & 42.59\tiny{$\pm$0.34} & 37.79\tiny{$\pm$0.78} & 37.24\tiny{$\pm$1.78} & 40.45\tiny{$\pm$0.89} & 36.36\tiny{$\pm$1.07} & 38.48\tiny{$\pm$1.05} \\
        XLM-R (es)  & 37.25\tiny{$\pm$2.33} & 39.38\tiny{$\pm$1.96} & 37.29\tiny{$\pm$1.12} & 39.25\tiny{$\pm$1.55} & 35.82\tiny{$\pm$1.01} & 38.98\tiny{$\pm$1.38} & 38.32\tiny{$\pm$1.47} & 39.51\tiny{$\pm$1.92} & 38.40\tiny{$\pm$0.87} & 35.73\tiny{$\pm$0.69} & 37.99\tiny{$\pm$1.51} \\
        \midrule
        \multicolumn{1}{l}{\textit{Zero-shot w/ adaptation}} \\
        \midrule
        XLM-R \texttt{+MLM} (en)   & 43.51\tiny{$\pm$1.69} & 38.13\tiny{$\pm$1.75} & 39.47\tiny{$\pm$1.19} & 52.44\tiny{$\pm$0.93} & 37.25\tiny{$\pm$2.60} & 46.21\tiny{$\pm$0.72} & 37.03\tiny{$\pm$3.28} & 61.78\tiny{$\pm$2.42} & 41.34\tiny{$\pm$0.61} & 39.82\tiny{$\pm$0.95} & 43.70\tiny{$\pm$1.83} \\
        XLM-R \texttt{+MLM} (es)  & 43.87\tiny{$\pm$0.14} & 40.05\tiny{$\pm$2.20} & 38.76\tiny{$\pm$0.08} & 52.27\tiny{$\pm$1.20} & 37.82\tiny{$\pm$1.59} & 44.17\tiny{$\pm$1.76} & \textbf{40.55\tiny{$\pm$1.07}} & \textbf{62.40\tiny{$\pm$1.44}} & 40.18\tiny{$\pm$0.95} & 38.45\tiny{$\pm$0.86} & 43.85\tiny{$\pm$1.30} \\
        \midrule
        \multicolumn{1}{l}{\textit{Translate-train}} \\
        \midrule
        XLM-R  & \textbf{50.00\tiny{$\pm$1.51}} & \textbf{51.42\tiny{$\pm$1.24}} & \textbf{42.45\tiny{$\pm$1.63}} & \textbf{58.89\tiny{$\pm$2.70}} & \textbf{43.20\tiny{$\pm$2.07}} & \textbf{55.33\tiny{$\pm$1.12}} & 36.01\tiny{$\pm$0.74} & 59.91\tiny{$\pm$0.20} & \textbf{52.00\tiny{$\pm$0.27}} & \textbf{42.04\tiny{$\pm$1.81}} & \textbf{49.12\tiny{$\pm$1.52}} \\
        \midrule
        \multicolumn{1}{l}{\textit{Translate-test}} \\
        \midrule
        XLM-R  & 39.73\tiny{$\pm$0.27} & 40.40\tiny{$\pm$0.13} & 34.71\tiny{$\pm$0.73} & 46.62\tiny{$\pm$2.29} & 38.00\tiny{$\pm$0.48} & 41.37\tiny{$\pm$0.16} & 35.29\tiny{$\pm$1.15} & 51.38\tiny{$\pm$1.24} & 39.51\tiny{$\pm$0.47} & 35.16\tiny{$\pm$0.97} & 40.22\tiny{$\pm$1.01} \\

    \bottomrule
    \end{tabular}
    \end{adjustbox}
    \caption{Results for zero-shot, translate-train, and translate-test averaged over 3 runs with different seeds. The majority baseline represents expected performance when predicting only the majority class of the test set. Random guessing would result in an accuracy of 33.33\%. Standard deviations in the Avg. column are calculated by taking the square root of the average variance of the languages in that row.}
    \label{tab:results}
\end{table*}

\paragraph{Pretrained Model}
We use XLM-R \cite{Conneau2020UnsupervisedCR} as the pretrained multilingual model in our experiments. The architecture of XLM-R is based on RoBERTa \cite{Liu2019roberta}, and it is trained using MLM on web-crawled data in 100 languages. It uses a shared vocabulary consisting of 250k subwords, created using SentencePiece \cite{kudo-richardson-2018-sentencepiece} tokenization. We use the \texttt{Base} version of XLM-R for our experiments.

\paragraph{Adaptation Methods}
To adapt XLM-R to the various target languages, we continue training with the MLM objective on monolingual text in the target language before finetuning. To keep a fair comparison with other approaches, we only use target data which was also used to train the translation models, which we describe in Section \ref{translation_models_section}. However, we note that one benefit of continued pretraining for adaptation is that it does not require parallel text, and could therefore benefit from text which could not be used for a translation-based approach. For continued pretraining, we use a batch size of 32 and a learning rate of 2e-5. We train for a total of 40 epochs. Each adapted model starts from the same version of XLM-R, and is adapted individually to each target language, which leads to a different model for each language. We denote models adapted with continued pretraining as \texttt{+MLM}.

\paragraph{Finetuning}
To finetune XLM-R, we follow the approach of \citet{Devlin2019} and use an additional linear layer. We train on either the English MNLI data or the machine-translated Spanish data, and we call the final models XLM-R (en) and XLM-R (es), respectively. Following \citet{hu2020xtreme}, we use a batch size of 32 and a learning rate of 2e-5. We train for a maximum of 5 epochs, and evaluate performance every 2500 steps on the XNLI development set.
We employ early stopping with a patience of 15 evaluation steps and use the best performing checkpoint for the final evaluation. All finetuning is done using the Huggingface Transformers library \cite{wolf-etal-2020-transformers} with up to two Nvidia V100 GPUs. Using \citet{lacoste2019quantifying}, we estimate total carbon emissions to be 75.6 kgCO$_2$eq.

\subsection{Translation-based Approaches}

We also experiment with two translation-based approaches, translate-train and translate-test, detailed below along with the translation model used.

\paragraph{Translation Models}
\label{translation_models_section}
For our translation-based approaches, we train two sets of translation models: one to translate from Spanish into the target language, and one in the opposite direction. We use transformer sequence-to-sequence models \cite{vaswani2017} with the hyperparameters proposed by \citet{guzman2019flores}. Parallel data used to train the translation models can be found in Table \ref{tab:parallel_data}. We employ the same model architecture for both translation directions, and we measure translation quality in terms of BLEU \cite{papineni-etal-2002-bleu} and ChrF \cite{popovic-2015-chrf}, cf. Table \ref{tab:translation_scores}. We use fairseq \cite{ott2019fairseq} to implement all translation models.\footnote{The code for translation models can be found at \url{https://github.com/AmericasNLP/americasnlp2021}}

\paragraph{Translate-train} For the translate-train approach, the Spanish training data provided by XNLI is translated into each target language. It is then used to finetune XLM-R for each language individually. Along with the training data, we also translate the Spanish development data, which is used for validation and early stopping. We discuss the effects of using a translated development set in Section \ref{early_stopping_app}. Notably, we find that the finetuning hyperparameters defined above do not reliably allow the model to converge for many of the target languages. To find suitable hyperparameters, we tune the batch size and learning rate by conducting a grid search over \{5e-6, 2e-5, 1e-4\} for the learning rate and \{32, 64, 128\} for the batch size. In order to select hyperparameters which work well across all languages, we evaluate each run using the average performance on the machine-translated Aymara and Guaraní development sets, as these languages have moderate and high ChrF scores, respectively. We find that decreasing the learning rate to 5e-6 and keeping the batch size at 32 yields the best performance. Other than the learning rate, we use the same approach as for zero-shot finetuning.

\paragraph{Translate-test} For the translate-test approach, we translate the test sets of each target language into Spanish. This allows us to apply the model finetuned on Spanish, XLM-R (es), to each test set. Additionally, a benefit of translate-test over translate-train and the adapted XLM-R models is that we only need to finetune once overall, as opposed to once per language. For evaluation, we use the checkpoint with the highest performance on the Spanish XNLI development set.

\begin{table*}[ht]
    \centering
    \small
    \setlength{\tabcolsep}{2.9pt}

    \begin{tabular}{l|ccccccccccc|c|c}
        \toprule
        & \textbf{FT} & \textbf{aym} & \textbf{bzd} & \textbf{cni} & \textbf{gn} & \textbf{hch} & \textbf{nah} & \textbf{oto} & \textbf{quy} & \textbf{shp} & \textbf{tar} & \textbf{Avg.} & \textbf{Avg.+P}\\
        \midrule
        Majority baseline & - & 33.33 & 33.33 & 33.33 & 33.33 & 33.33 & 33.47 & 33.42 & 33.33 & 33.33 & 33.33 & - & - \\
        \midrule
        \multicolumn{1}{l}{\textit{Zero-shot}} \\
        \midrule
         XLM-R (en) & 62.34 & 33.60 & 33.47 & 32.40 & 33.47 & 34.13 & 33.06 & 32.35 & 33.33 & 33.60 & 34.27 & 33.37& 38.48 \\
        XLM-R (es) & 62.26 & 34.13 & 34.80 & 35.33 & 35.33 & 34.53 & 33.60 & 33.16 & 33.07 & 36.80 & 35.73 & 34.65& 37.99 \\
        \midrule
        \multicolumn{1}{l}{\textit{Zero-shot w/ adaptation}} \\
        \midrule
        XLM-R \texttt{+MLM} (en) & - & 37.07 & 32.80 & 33.07 & 42.40 & 33.73 & 34.55 & 33.96 & 44.40 & 35.33 & 34.80 & 36.21& 43.70 \\
        XLM-R \texttt{+MLM} (es) & - & 36.27 & 34.80 & 33.73 & 41.73 & 34.00 & 35.37 & 32.89 & 47.87 & 35.60 & 34.67 & 36.69& 43.85 \\
        \midrule
        \multicolumn{1}{l}{\textit{Translate-train}} \\
        \midrule
        XLM-R & - & \textbf{44.93} & \textbf{43.73} & \textbf{43.47} & \textbf{47.60} & \textbf{43.07} & \textbf{45.80} & \textbf{35.83} & \textbf{52.13} & \textbf{46.27} & \textbf{39.47} & \textbf{44.23} & 49.12  \\
        \midrule
        \multicolumn{1}{l}{\textit{Translate-test}} \\
        \midrule
        XLM-R & - & 36.53 & 42.67 & 37.33 & 43.60 & 38.53 & 43.22 & 34.22 & 48.13 & 42.67 & 34.67 & 40.16 & 40.22 \\

    \bottomrule
    \end{tabular}

    \caption{ Hypothesis-only results. The \textit{Avg.} column represents the average of the hypothesis-only results, while the \textit{Avg.+P} column, taken from Table \ref{tab:results}, represents the average of the languages when using both the premise and hypothesis.}
    \label{tab:hyp_only_results}
\end{table*}

\section{Results and Discussion}

\paragraph{Zero-shot Models} We present our results in Table \ref{tab:results}. Results for the development set are presented in Table \ref{tab:dev_results}. Zero-shot performance is low for all 10 languages, with an average accuracy of 38.48\% and 37.99\% for the English and Spanish model, respectively. However, in all cases the performance is higher than the majority baseline. As shown in Table \ref{tab:xnli} in the appendix, the same models achieve an average of 74.20\% and 75.35\% accuracy respectively, when evaluated on the 15 XNLI languages. 

 Interestingly, even though code-switching with Spanish is encountered in many target languages, finetuning on Spanish labeled data on average slightly underperforms the model trained on English, however performance is better for 3 of the languages. The English model achieves a highest accuracy of 42.59\%, when evaluated on Nahuatl, while the Spanish model achieves a highest accuracy of 39.51\%, when evaluated on Quechua. The lowest performance is achieved when evaluating on Aymara and Rarámuri, for the English and Spanish model, respectively.

We find that model adaptation via continued pretraining improves both models, with an average gain of 5.22 percentage points for English and 5.86 percentage points for Spanish. Notably, continued pretraining increases performance for Quechua by 24.53 percentage points when finetuning on English, and 22.89 points when finetuning on Spanish. Performance decreases for Bribri and Otomí when finetuning on English, however performance for all languages improves when using Spanish.

\paragraph{Translate-test} Performance of the translate-test model improves over both zero-shot baselines. We see the largest increase in performance for Guaraní and Quechua, with gains of 7.16 and, respectively, 11.87 points over the best performing zero-shot model without adaptation. Considering the translation metrics in Table \ref{tab:translation_scores}, models for Guaraní and Quechua achieve the two highest scores for both metrics. On average, translate-test does worse when compared to the adapted zero-shot models, and in all but two cases, both adapted models perform better than translate-test. We hypothesize that translate-test is more sensitive to noise in the translated data; sentences may lose too much of their original content, preventing correct classification.

\paragraph{Translate-train} The most surprising result is that of translate-train, which considerably outperforms the performance of translate-test for all languages, and outperforms the zero-shot models for all but two languages. Compared to the best non-adapted zero-shot model, the largest performance gain is 20.40 points for Quechua. For the language with the lowest performance, Otomí, translate-train performs 2.32 points worse than zero-shot; however, it still outperforms translate-test. When averaged across all languages, translate-train outperforms the English zero-shot model by 10.64 points, and translate-test by 8.9 points.
It is important to note that the translation performance from Spanish to each target language is not particularly high: when considering ChrF scores, the highest is 0.33, and the highest BLEU score is 3.26. Performance of both translation-based models is correlated with ChrF scores, with a Pearson correlation coefficient of 0.82 and 0.83 for translate-train and translate-test. Correlations are not as strong for BLEU, with coefficients of 0.37 and 0.59.

The sizable difference in performance between translate-train and the other methods suggests that translation-based approaches may be a valuable asset for cross-lingual transfer, especially for low-resource languages. While the largest downsides to this approach are the requirement for parallel data and the need for multiple models, the potential performance gain over other approaches may prove worthwhile. Additionally, we believe that the performance of both translation-based approaches would improve given a stronger translation system, and future work detailing the necessary level of translation quality for the best performance would offer great practical usefulness for NLP applications for low-resource languages.

\section{Analysis}
\label{analysis}

\subsection{Hypothesis-only Models }
As shown by \citet{Gururangan2018AnnotationAI}, SNLI and MNLI
-- the datasets AmericasNLI is based on --
contain artifacts created during the annotation process which models exploit to artificially inflate performance. To analyze whether similar artifacts exist in AmericasNLI and if they can also be exploited, we train and evaluate models using only the hypothesis, and present results in Table \ref{tab:hyp_only_results}. We can see that the average performance across languages is better than chance for all models except for XLM-R without adaptation. Translate-train obtains the highest result with $44.23\%$ accuracy, and as shown in Table \ref{tab:hyp_only_differences}, hypothesis-only performance of translate-test is higher than standard performance for 5 languages. Thus, as with SNLI and MNLI, artifacts in the hypotheses can be used to predict, to some extent, the correct labels. However all but 1 zero-shot and translate-train models perform better in the standard setting, indicating that the models are learning something beyond just exploiting artifacts in the hypotheses, even 
with the additional challenge of unseen languages.

\subsection{Case Study: Human Evaluation}

Following \citet{conneau2018xnli}, AmericasNLI was created by translating sentences individually, in order to prevent additional context being added into the hypotheses. However, this strategy may break the original semantic relationship between the premise and the hypothesis. Furthermore, for some examples the logical relationship may be dependent on context or subtext which can be lost through translation, or simply not make sense in the target language. To verify the validity of the labels of AmericasNLI, we conduct a human evaluation experiment, focusing on examples translated to Bribri. We create a balanced, random sample of 450 examples taken from the Bribri development set. An annotator familiar with the task was then asked to classify the pairs of sentences. For comparison, we also annotate parallel examples taken from the English and Spanish development sets. For Bribri, we recover the original XNLI label for 76.44\% of examples. For English and Spanish, we achieve 81.78\% and 71.56\% accuracy, respectively. Due to the relatively small differences in performance across languages, we conclude that translation to Bribri has a minimal effect on the semantic relationship between the premise and the hypothesis.

\section{Limitations and Future Work} 

While the case study above provides strong evidence for the validity of our Bribri examples, we cannot currently generalize this claim to the remaining languages. For future work, we plan on extending our human evaluation to more languages and provide a more detailed analysis. 
 
Additionally, due to the limited availability of annotators and the difficulties of translation for languages that are less frequently studied, the size of the AmericasNLI test set is relatively small. As such, care must be taken to carefully evaluate conclusions drawn using the dataset; following \citet{Card2020WithLP} we present a power analysis of our results in Section \ref{sec:power_analysis}. Future work expanding the dataset size will help create a stronger baseline. Furthermore, while we do not make any model-specific assumptions in our experiments, our results are based on only one pretrained model and adaptation method. Methods using vocabulary extension or adapters may offer additional improvements. Similarly, other pretrained models could perform differently, depending on, e.g.,
the model size or the set of languages in their pretraining data. In Table \ref{tab:xlmr_large_results}, we present results using XLM-R Large, and find that, while the relationship between the approaches differs from the main experiments, the overall highest average performance is still achieved by the translate-train approach with XLM-R Base. We provide a longer discussion in Section \ref{sec:large}.

\section{Conclusion}
 To better understand the zero-shot abilities of pretrained multilingual models for semantic tasks in unseen languages, we present AmericasNLI, a parallel NLI dataset covering 10 low-resource languages indigenous to the Americas.
 We conduct experiments with XLM-R, and find that the model's zero-shot performance, while better than a majority baseline, is poor. However, it can be improved by model adaptation via continued pretraining. Additionally, we find that translation-based approaches outperform a zero-shot approach, which is surprising given the low quality of the employed translation systems. We hope that this work will not only spur further research into improving model adaptation to unseen languages, but also motivate the creation of more resources for languages not frequently studied by the NLP community.
 
\section*{Ethics Statement} In this work, we present a new dataset created through the translation of an existing resource, XNLI \cite{conneau2018xnli}. While this allows for results that are directly comparable, it also means that this dataset inherits any biases and flaws which are contained in the previous dataset. Furthermore, research involving languages spoken by Indigenous communities raises ethical concerns regarding the exploitation of these languages and communities: it is crucial that members of the community are able to directly benefit from the research. Translation for AmericasNLI was done by either paper authors or translators who were compensated at a rate based on the average rate for translation and the minimum wage in their country of residence. Additionally, many authors are members of, and/or have a record of close work with communities who speak a language contained in AmericasNLI.

\section*{Acknowledgments}
We thank the following people for their work on the translations:
Francisco Morales for Bribri, Feliciano Torres Ríos for Asháninka, Perla Alvarez Britez for Guaraní, Silvino González de la Crúz for Wixarika, Giovany Martínez Sebastián, Pedro Kapoltitan, and José Antonio for Nahuatl, José Mateo Lino Cajero Velázquez for Otomí, Liz Chávez for Shipibo-Konibo, and María del Cármen Sotelo Holguín for Rarámuri. We would also like to thank Dallas Card for his help with power analysis. This work would not have been possible without the financial support of Facebook AI Research, Microsoft Research, Google Research, the Institute of Computational Linguistics at the University of Zurich, the NAACL Emerging Regions Fund, Comunidad Elotl, and Snorkel AI.

\bibliography{anthology,custom}
\bibliographystyle{acl_natbib}

\clearpage

\appendix

\section{Geographic Distribution of the AmericasNLI Languages}

\renewcommand{\thefigure}{A.\arabic{figure}}
\begin{minipage}{\textwidth}

    \centering
    \begin{minipage}{.4\textwidth}
    \centering
    \includegraphics[scale=0.75]{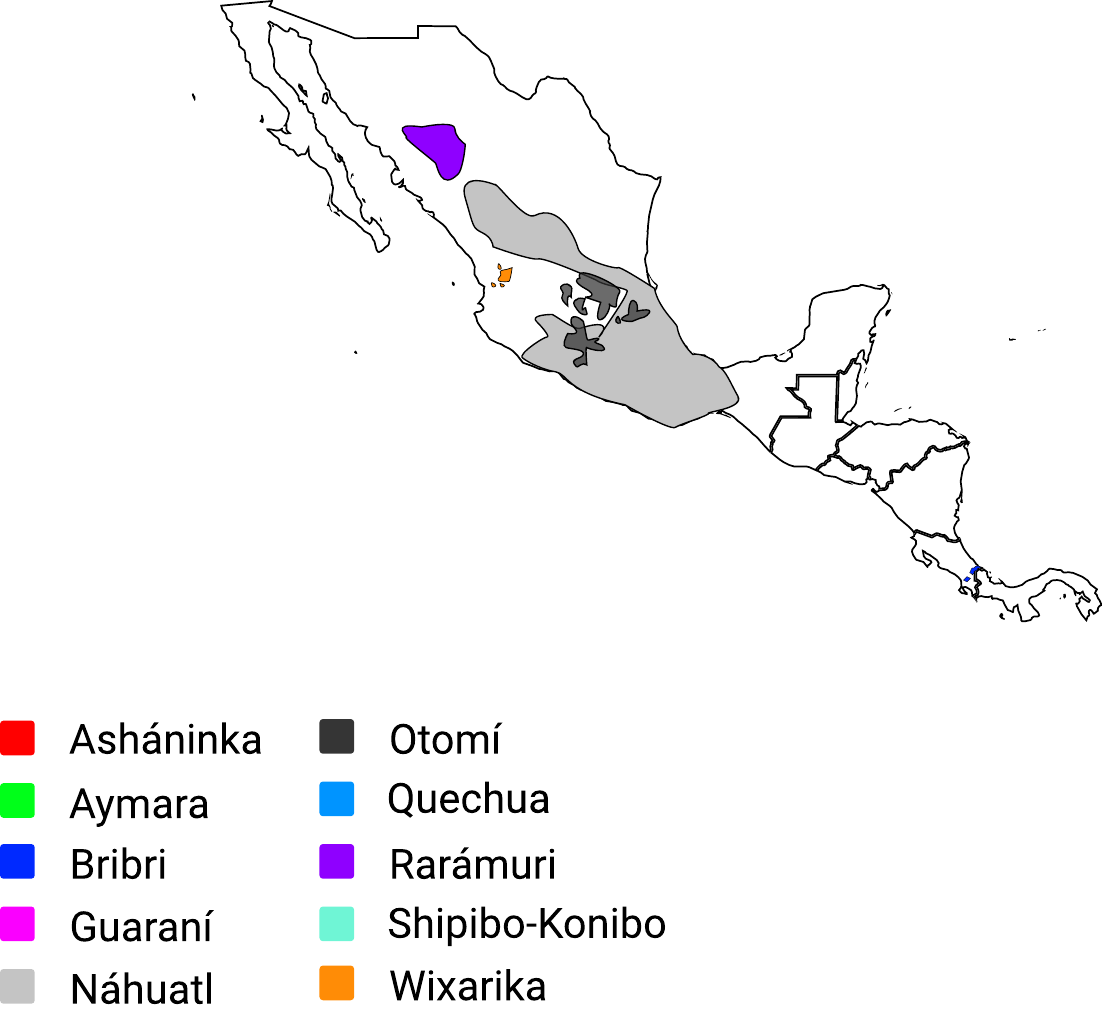}
    \end{minipage}%
    \begin{minipage}{.68\textwidth}
    \centering
    \includegraphics[scale=0.55]{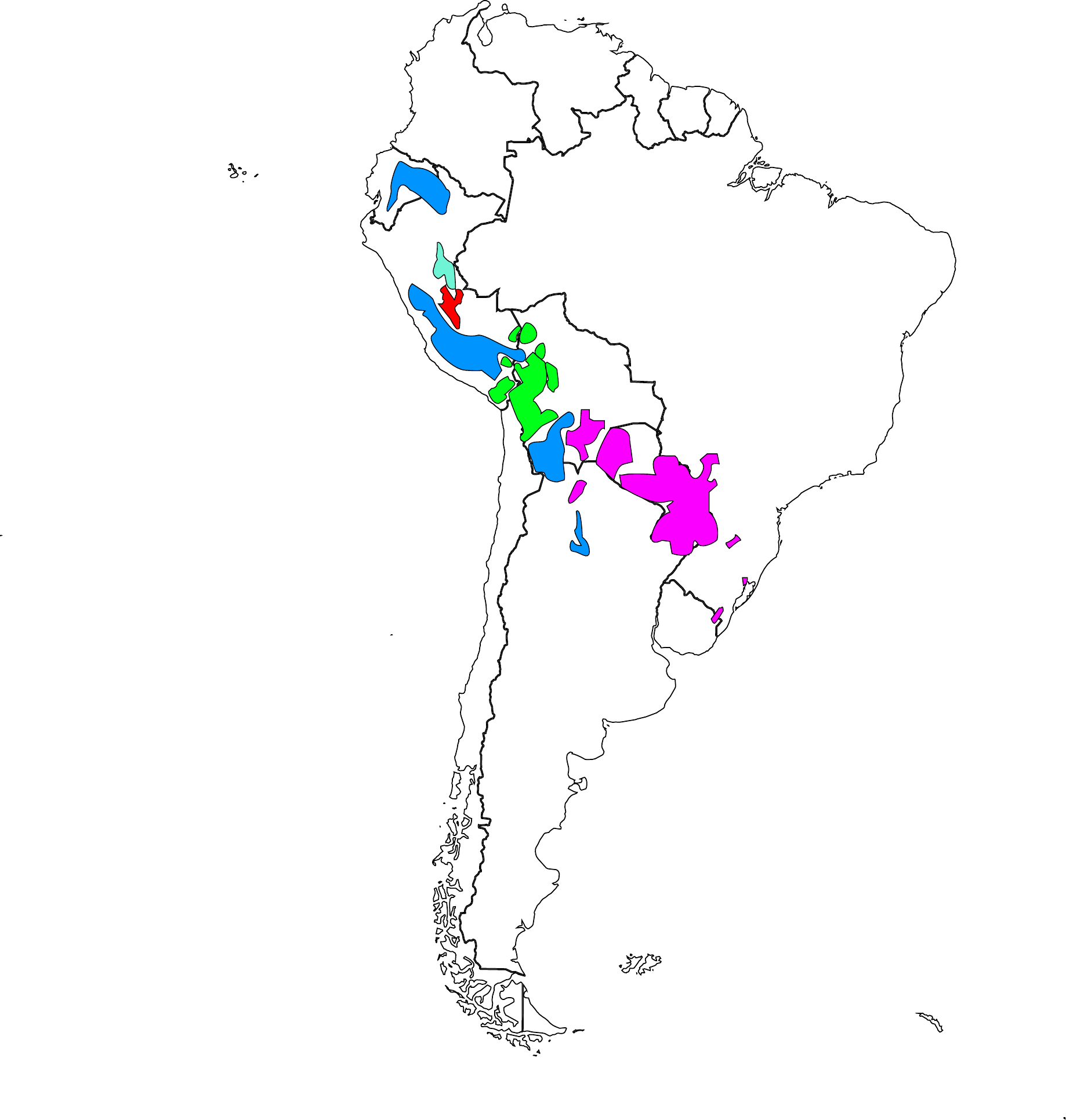}
    \end{minipage}
    \captionof{figure}{Maps of Central and South America presenting an approximate distribution of where each Indigenous language contained in AmericasNLI is spoken. Please note that this map is hand-drawn and largely an estimate: some regions may not be included, and borders of included regions may not be completely accurate.}

\end{minipage}

\section{Sources of Parallel Data}
\text{ } \\
\renewcommand{\thetable}{B.\arabic{table}}
\begin{minipage}{\textwidth}
    \centering
    \footnotesize
    \setlength{\tabcolsep}{1pt}
    \small

    \begin{tabular}{clr}
        \toprule
        \textbf{Lang.} & \multicolumn{1}{c}{\textbf{Source(s)}} & \textbf{Sent.} \\
        \midrule
        {aym} & \citet{tiedemann-2012-parallel} & 6,531 \\
        \midrule
        \multirow{4}{*}{bzd} & \citet{feldman-coto-solano-2020-neural}; \citet{margery}; & \multirow{4}{*}{7,508} \\
         &\citet{gramaticacarla}; \citet{cursobribri}; & \\
         & \citet{cursoCarla}; & \\
         &\citet{itte}; \citet{corpusSofia} & \\
         \midrule
        {cni} & \citet{cushimariano:prel:08}& 3,883 \\
        \midrule
        {gn} & \citet{chiruzzo-etal-2020-development} &  26,032\\
        \midrule
        {hch} & \citet{mager-wixarika} & 8,966 \\
        \midrule
        {nah} & \citet{gutierrez-vasques-etal-2016-axolotl} & 16,145 \\
        \midrule
        {oto} & \url{https://tsunkua.elotl.mx} & 4,889 \\
        \midrule
        {quy} & \citet{agic-vulic-2019-jw300} & 125,008 \\
        \midrule
        \multirow{2}{*}{shp} & \citet{galarreta-etal-2017-corpus}; \citet{james-1993-diccionario}; & \multirow{2}{*}{14,592} \\
         & \citet{gomez-etal-2019-continuous} & \\
        \midrule
        \multirow{2}{*}{tar} & \citet{brambila_1976};
        & \multirow{2}{*}{14,720}\\
        & \url{github.com/pywirrarika/tar\_par} & \\

    \bottomrule
    \end{tabular}
    \captionsetup{justification=justified}\addtocounter{table}{-5}
    \captionof{table}{Parallel data used for our translation models.}
    \label{tab:parallel_data}
\end{minipage}
\text{ } \\ \\
\twocolumn
\section{Additional Information for AmericasNLI Languages}

\subsection{Aymara} 
A rare linguistic phenomenon found in Aymara is vowel elision, a deletion of certain vowel sounds triggered by complex phonological, morphological, and syntactic factors.  

\subsection{Asháninka}
While Asháninka in a strict sense refers to the linguistic varieties spoken in Ene, Tambo and Bajo Perené rivers, the name is also used to talk about the following nearby and closely-related Asheninka varieties: Alto Perené, Pichis, Pajonal, Ucayali-Yurua, and Apurucayali. Although Asháninka is the most widely spoken Amazonian language in Peru, certain varieties, such as Alto Perené, are highly endangered.

The verb is the most morphologically complex word class, with a rich repertoire of aspectual and modal categories. The language lacks case, except for one locative suffix, so the grammatical relations of subject and object are indexed as affixes on the verb itself. Other notable linguistic features of the language include obligatory marking of a realis/irrealis distinction on the verb, a rich system of applicative suffixes, serial verb constructions, and a pragmatically conditioned split intransitivity. 

\subsection{Bribri}
As previously noted, Bribri is a vulnerable language, and there are few settings where the language is written or used in official functions. The language does not have official status and it is not the main medium of instruction of Bribri children, but it is offered as a class in primary and secondary schools. Bribri features fusional morphology and an ergative-absolutive case system. Bribri grammar also includes phenomena like head-internal relative clauses, directional verbs and numerical classifiers \cite{gramaticacarla}.

\subsection{Guaraní}
While the first written record dates to 1591, Guaraní usage in text continued until the Paraguay-Triple Alliance War (1864-1870) and declined thereafter. From the 1920s on, Guaraní has slowly re-emerged and received renewed focus. In 1992, Guaraní was the first American language declared an official language of a country, followed by a surge of local, national, and international recognition in the early 21st century.\footnote{\url{https://es.wikipedia.org/wiki/Idioma_guarani}} 
\subsection{Nahuatl} Nahuatl is spoken in 17 different states of Mexico. In Nahuatl, different roots with or without affixes are combined to form new words. The suffixes that are added to a word modify the meaning of the original word \cite{sullivan1976compendio}, and 18 prepositions stand out based on postpositions of names and adjectives \cite{simeon1977diccionario}.

\subsection{Otomí}
The various regional self-denominations of Otomí include \textit{ñähñu} or \textit{ñähño}, \textit{hñähñu, ñuju, ñoju, yühu, hnähño, ñühú, ñanhú, ñöthó, ñható} and \textit{hñothó} \citep{NormaOto}. Many words are homophonous to Spanish \cite{cajero1998, Mateo2009}. When speaking \textit{ñuhmu}, pronunciation is elongated, especially on the last syllable. The alphabet is composed of 19 consonants, 12 vowel phonemes.

\subsection{Rarámuri}
Rarámuri 
is mainly spoken in the state of Chihuahua. There are five variants of Rarámuri.
\subsection{Shipibo-Konibo}
Shipibo-Konibo is a language with agglutinative processes, a majority of which are suffixes. However, clitics are also used, and are a widespread element in Panoan literature \cite{valenzuela-2003-transitivity}.
\subsection{Wixarika}
The four variants of Wixarika are the Northern, Southern, Eastern, and Western variants \cite{inegi2008catalogo}. It is spoken mainly in the three Mexican states of Jalisco, Nayari, and Durango. Features of Wixarika include head-marking \cite{nichols1986head}, a head-final structure \cite{greenberg1963universals}, nominal incorporation, argumentative marks, inflected adpositions, possession marks, as well as instrumental and directional affixes \cite{iturrioz2008gramatica}.

\subsection{Summary of Language Information}
\renewcommand{\thetable}{C.\arabic{table}}
\setcounter{table}{0}
\vspace{5mm}
\begin{minipage}{\textwidth}

\centering
\begin{adjustbox}{width=\textwidth}
\begin{tabular}{lllll}
        \toprule
        \textbf{Language} & \textbf{Language Family} & \textbf{Countries Spoken} & \textbf{Number of Speakers} & \textbf{Word Order} \\
        \midrule
        aym & Aymaran & Bolivia, Chile, Peru & 2m & SOV \\
    bzd & Chibchan & Costa Rica & 7k & SOV \\
    cni  & Arawak & Peru & 73k & VSO \\
    gn & Tupi-Guarani & Paraguay, Brazil, Argentina, Bolivia & 6-10m & SVO \\
    hch & Uto-Aztecan & Mexico & 47k & SOV \\
    nah & Uto-Aztecan & Mexico & 1.5m & SVO/VSO/SOV \\
    oto & Oto-Manguean & Mexico & 307k & SVO \\
    quy & Quechuan & Peru & 8-10m & SOV \\
    shp & Panoan & Peru & 35k & SOV \\
    tar & Uto-Aztecan & Mexico & 89k & SOV \\
        \bottomrule
\end{tabular}
\end{adjustbox}
\captionof{table}{Summary of the 10 languages in AmericasNLI. }
\label{tab:language_summary}
\end{minipage}

\section{Dataset Information}
\renewcommand{\thetable}{D.\arabic{table}}
\setcounter{table}{0}
\subsection{Power Analysis}
\label{sec:power_analysis}
\vspace{3cm}
\begin{minipage}{\textwidth}

\centering
\begin{adjustbox}{width=0.9\columnwidth}
\begin{tabular}{cccccl}
        \toprule
        $p1$ Model & \textbf{$p1$} & \textbf{$p2$} & \textbf{Lower Bound Power} & \textbf{Upper Bound Power} & $p2$ Model \\
        \midrule
         \multirow{6}{*}{Random Baseline} & \multirow{6}{*}{33.33} & 38.48 & 40.33 & 100 & Zero-shot (en) \\
        && 37.99 & 35.80 & 100 & Zero-shot (es) \\
        && 43.70 & \textbf{91.38} & 100 & Zero-shot +MLM (en) \\
        && 43.85 & \textbf{91.52} & 100 & Zero-shot +MLM (es) \\
        && 49.12 & \textbf{99.82} & 100 & Translate-train \\
        && 40.22 & 61.85 & 100 & Translate-test \\
        \midrule
        \multirow{4}{*}{Zero-shot Baseline} & \multirow{4}{*}{38.48} & 43.70 & 33.66 & 100 & Zero-shot +MLM (en) \\
        && 43.85 & 35.33 & 100 & Zero-shot +MLM (es) \\
        && 49.12 & \textbf{87.10} & 100 & Translate-train \\
        && 40.22 & 7.13 & 99.07 & Translate-test \\
        \midrule
       Adaptation Baseline & 43.85 & 49.12 & 31.29 & 100 & Translate-train \\
        \bottomrule
\end{tabular}
\end{adjustbox}
\captionof{table}{Here, we use the simulation approach of \citet{Card2020WithLP} to calculate upper and lower bounds for the power of our experiments. We use the average accuracies for each approach, and set $n=750, \alpha=0.05, r=10,000$, and bold experiments with well-powered lower bounds.}

\label{tab:post_hoc_power_analysis}
\end{minipage}

\clearpage

\subsection{Dataset Statistics}
\text{ } \\ \\ 
\setcounter{table}{2}
\begin{minipage}[c]{\textwidth}
    \centering
    \small

    \begin{tabular}{ccccc|c}
        \toprule
          \textbf{Language} & \textbf{Split} & \textbf{Entailment} & \textbf{Contradiction} & \textbf{Neutral} & \textbf{Majority Baseline} \\
          \midrule
        \multirow{2}{*}{aym} & Test & 250 & 250 & 250 & 0.333 \\
         & Dev & 248 & 248 & 247 & 0.334 \\
         \midrule
        \multirow{2}{*}{bzd} & Test & 250 & 250 & 250 & 0.333 \\
        & Dev & 248 & 248 & 247 & 0.334 \\
        \midrule
        \multirow{2}{*}{cni} & Test & 250 & 250 & 250 & 0.333 \\
        & Dev & 220 & 220 & 218 & 0.334 \\
        \midrule
        \multirow{2}{*}{gn} & Test & 250 & 250 & 250 & 0.333 \\
        & Dev & 248 & 248 & 247 & 0.334 \\
        \midrule
        \multirow{2}{*}{hch} & Test & 250 & 250 & 250 & 0.333 \\
        & Dev & 248 & 248 & 247 & 0.334 \\
        \midrule
        \multirow{2}{*}{nah} & Test & 246 & 245 & 247 & 0.335 \\
         & Dev & 193 & 195 & 197 & 0.337 \\
        \midrule
        \multirow{2}{*}{oto} & Test & 249 & 249 & 250 & 0.334 \\
        & Dev & 78 & 75 & 69 & 0.351 \\
        \midrule
        \multirow{2}{*}{quy} & Test & 250 & 250 & 250 & 0.333 \\
        & Dev & 248 & 248 & 247 & 0.334 \\
        \midrule
        \multirow{2}{*}{shp} & Test & 250 & 250 & 250 & 0.333 \\
        & Dev & 248 & 248 & 247 & 0.334 \\
        \midrule
        \multirow{2}{*}{tar} & Test & 250 & 250 & 250 & 0.333 \\
        & Dev & 248 & 248 & 247 & 0.334 \\

    \bottomrule
    \end{tabular}
   \captionsetup{justification=justified}\addtocounter{table}{-1}

    \captionof{table}{Distribution of labels in the test and development sets, per language.}
\end{minipage}
\text{ } \\ \\

\section{Detailed Results}
\text{ } \\ \\
\renewcommand{\thetable}{E.\arabic{table}}
\setcounter{table}{0}
\begin{minipage}{\textwidth}
    \centering
    \small
    \setlength{\tabcolsep}{4pt}

    \begin{tabular}{l|ccccccccccc |c}
        \toprule
        & \textbf{FT} & \textbf{aym} & \textbf{bzd} & \textbf{cni} & \textbf{gn} & \textbf{hch} & \textbf{nah} & \textbf{oto} & \textbf{quy} & \textbf{shp} & \textbf{tar} & \textbf{Avg.}\\
        \midrule
        Majority baseline & - & 33.40 & 33.40 & 33.40 & 33.40 & 33.40 & 33.70 & 35.10 & 33.40 & 33.40 & 33.40 & - \\
        \midrule
        \multicolumn{1}{l}{\textit{Zero-shot}} \\
        \midrule
         XLM-R (en) & 84.55 & 38.45 & 41.59 & 40.07 & 40.74 & 37.82 & 39.50 & 43.84 & 38.67 & 43.56 & 36.03 & 40.03 \\
        XLM-R (es) & 80.77 & 37.73 & 39.70 & 37.59 & 40.06 & 36.74 & 37.88 & 39.94 & 38.54 & 38.18 & 35.89 & 38.23 \\
        \midrule
        \multicolumn{1}{l}{\textit{Zero-shot w/ adaptation}} \\
        \midrule
        XLM-R \texttt{+MLM} (en) & - & 41.77 & 39.57 & 40.93 & 52.40 & 41.01 & 43.25 & 37.24 & 62.27 & 44.86 & 39.30 & 44.26 \\
        XLM-R \texttt{+MLM} (es) & - & 45.26 & 42.22 & 40.53 & 53.52 & 38.40 & 42.41 & 40.24 & 55.00 & 40.11 & 45.89 & 44.36 \\
        \midrule
        \multicolumn{1}{l}{\textit{Translate-train}} \\
        \midrule
        XLM-R & - & 53.61 & 49.98 & 45.49 & 61.28 & 42.22 & 53.80 & 41.44 & 58.62 & 53.10 & 43.01 & 50.25 \\
        \midrule
        \multicolumn{1}{l}{\textit{Translate-test}} \\
        \midrule
        XLM-R & - & 37.73 & 39.70 & 37.59 & 40.06 & 36.74 & 37.88 & 39.94 & 38.54 & 38.18 & 35.89 & 38.23 \\

    \bottomrule
    \end{tabular}

    \captionof{table}{Development set results for zero-shot, translate-train, and translate-test. \textit{FT} represents the XNLI development set performance for the finetuning language and is not included in the average. The majority baseline represents expected performance when predicting only the majority class of the development set. Random guessing would result in an accuracy of 33.33\%.}
    \label{tab:dev_results}
\end{minipage}
\clearpage
\text{ } \\ \\ \\

\begin{minipage}{\textwidth}
    \centering
    \small
    \setlength{\tabcolsep}{4pt}
     \begin{tabular}{l|ccccccccccc|c}
        \toprule
        & \textbf{FT} & \textbf{aym} & \textbf{bzd} & \textbf{cni} & \textbf{gn} & \textbf{hch} & \textbf{nah} & \textbf{oto} & \textbf{quy} & \textbf{shp} & \textbf{tar} & \textbf{Avg.}\\
        \midrule
        \multicolumn{1}{l}{\textit{Zero-shot}} \\
        \midrule
         XLM-R (en) & -22.21 & -2.53 & -6.18 & -5.51 & -6.00 & -3.07 & -9.53 & -5.44 & -3.91 & -6.85 & -2.09 & -5.11 \\
        XLM-R (es) & -18.51 & -3.12 & -4.58 & -1.96 & -3.92 & -1.29 & -5.38 & -5.16 & -6.44 & -1.60 & 0.00 & -3.35 \\
        \midrule
        \multicolumn{1}{l}{\textit{Zero-shot w/ adaptation}} \\
        \midrule
        XLM-R \texttt{+MLM} (en) & - & -6.44 & -5.33 & -6.40 & -10.04 & -3.52 & -11.66 & -3.07 & -17.38 & -6.01 & -5.02 & -7.49 \\
        XLM-R \texttt{+MLM} (es) & -   & -7.60 & -5.25 & -5.03 & -10.54 & -3.82 & -8.80 & -7.66 & -14.53 & -4.58 & -3.78 & -7.16 \\
        \midrule
        \multicolumn{1}{l}{\textit{Translate-train}} \\
        \midrule
        XLM-R & - & -5.07 & -7.69 & 1.02 & -11.29 & -0.13 & -9.52 & -0.18 & -7.78 & -5.73 & -2.57 & -4.89 \\
        \midrule
        \multicolumn{1}{l}{\textit{Translate-test}} \\
        \midrule
        XLM-R & - & -3.20 & 2.27 & 2.62 & -3.02 & 0.53 & 1.85 & -1.07 & -3.25 & 3.16 & -0.49 & -0.06 \\
    \bottomrule
    \end{tabular}

    \captionof{table}{Differences between hypothesis-only and standard results on the test set of AmericasNLI.}
    \label{tab:hyp_only_differences}
\end{minipage}

\text{ } \\ \\

\begin{minipage}{0.98\textwidth}
    \centering
    \setlength{\tabcolsep}{2pt}
    \small

    \begin{tabular}{c|ccccccccccccccc|c}
        \toprule
        \textbf{Source} & \textbf{ar} & \textbf{bg} & \textbf{de} & \textbf{el} & \textbf{en} & \textbf{es} & \textbf{fr} & \textbf{hi} & \textbf{ru} & \textbf{sw} & \textbf{th} & \textbf{tr} & \textbf{ur} & \textbf{vi} & \textbf{zh} & \textbf{Avg.} \\
        \midrule
en & 71.96 & 77.65 & 76.62 & 75.84 & \underline{84.55} & 78.74 & 78.00 & 70.02 & 76.04 & 64.41 & 72.04 & 72.54 & 66.28 & 74.38 & 73.97 & 74.20 \\
es & 73.49 & 78.71 & 77.59 & 77.05 & 83.36 & \underline{80.77} & 78.83 & 72.25 & 77.10 & 64.60 & 73.32 & 73.78 & 68.44 & 75.82 & 75.16 & 75.35 \\

    \bottomrule
    \end{tabular}

    \captionof{table}{Results of zero-shot models on the test set of XNLI. Scores are underlined when the same language used for training is used for evaluation as well.}
    \label{tab:xnli}
\end{minipage}

\text{ } \\ \\ \\
\section{Additional Results}
\text{ } \\ \\
\renewcommand{\thetable}{F.\arabic{table}}
\setcounter{table}{0}
\begin{minipage}{\textwidth}

\centering
\begin{adjustbox}{width=\textwidth}
\begin{tabular}{cl|cccccccccc|c}
        \toprule
        \textbf{Source} & \textbf{Model} & \textbf{aym} & \textbf{bzd} & \textbf{cni} & \textbf{gn} & \textbf{hch} & \textbf{nah} & \textbf{oto} & \textbf{quy} & \textbf{shp} & \textbf{tar} & \textbf{Avg.} \\
        \midrule
        \multirow{3}{*}{en} & Zero-Shot & 36.00 & 39.20 & 37.20 & 40.67 & 36.80 & 42.28 & 36.90 & 35.73 & 40.67 & 36.27 & 38.17 \\
        & Z-S +MLM & 41.60 & 36.53 & 40.80 & 51.47 & 39.87 & 46.48 & 37.83 & 64.53 & 40.67 & 40.67 & \textbf{44.05} \\
        & Z-S $+MLM_{AUG}$ & 45.07 & 38.67 & 41.47 & 52.93 & 38.53 & 46.48 & 33.42 & 62.00 & 39.73 & 40.27 & 43.86 \\
        \midrule
        \multirow{3}{*}{es} & Zero-Shot & 37.87 & 41.60 & 37.87 & 39.47 & 36.27 & 39.57 & 39.04 & 40.93 & 38.27 & 35.33 & 38.62 \\
        & Z-S +MLM & 43.87 & 37.60 & 38.80 & 52.27 & 36.00 & 45.12 & 41.58 & 60.80 & 41.20 & 38.80 & 43.60 \\
        & Z-S $+MLM_{AUG}$ & 45.20 & 38.67 & 39.33 & 54.27 & 37.07 & 44.99 & 42.65 & 62.67 & 37.20 & 38.67 & \textbf{44.07}
        \\
        \midrule
        \multirow{3}{*}{$-$}& Translate-Train & 49.33 & 52.00 & 42.80 & 55.87 & 41.07 & 54.07 & 36.50 & 59.87 & 52.00 & 43.73 & 48.72 \\
        & T-T +MLM & 50.93 & 51.20 & 42.27 & 61.60 & 44.93 & 56.10 & 35.16 & 63.47 & 50.00 & 44.13 & \textbf{49.98} \\
        & T-T $+MLM_{AUG}$ & 51.07 & 51.87 & 44.53 & 61.07 & 46.27 & 53.39 & 35.96 & 61.07 & 52.67 & 40.67 & 49.86 \\
        \bottomrule
\end{tabular}
\end{adjustbox}
\captionof{table}{Results from models adapted with augmented data before finetuning. Zero-shot, zero-shot +MLM, and translate-train results are taken from the main experiments, however we only take results from the run corresponding to the same random seed as the newly trained models.}
\label{tab:mlm_aug_results}
\end{minipage}
\clearpage

\subsection{Early Stopping}
\label{early_stopping_app}

While early stopping is vital for machine learning, in the case of zero-shot learning hand-labeled development sets in the target language are often assumed to be unavailable \cite{kann-etal-2019-towards}. Thus, in our main experiments we use either a machine-translated development set or one from a high-resource language. In both cases, performance on the development set is an imperfect signal for how the model will ultimately perform. To explore how this affects final performance, we present the difference in results for translate-train models when an oracle translation is used for early stopping in Table \ref{tab:oracle_translate_train}. We find that performance is 2.34 points higher on average, with a maximum difference of 7.28 points for Asháninka, suggesting that creating ways to better approximate a development set may lead to higher performance.

\begin{table}[h]
    \centering
    \setlength{\tabcolsep}{1.5pt}

    \begin{adjustbox}{width=\columnwidth}

    \begin{tabular}{cccccccccc|c}
        \toprule
        \textbf{aym} & \textbf{bzd} & \textbf{cni} & \textbf{gn} & \textbf{hch} & \textbf{nah} & \textbf{oto} & \textbf{quy} & \textbf{shp} & \textbf{tar} & \textbf{Avg.} \\
        \midrule
         2.13 & 0.98 & 7.28 & 0.58 & 0.53 & 2.12 & 3.03 & 1.42 & 0.93 & 4.36 & 2.34 \\

    \bottomrule
    \end{tabular}
    \end{adjustbox}
    \caption{Difference between translate-train results obtained using the oracle development set and the translated development set for early stopping.}
    \label{tab:oracle_translate_train}
\end{table}

\subsection{Data Augmentation with Translated Data}
Due to the success of translate-train, we also investigate if we can improve performance further by creating data for language adaptation (+MLM) through translation. To do so, we create a random sample of sentences taken from Spanish Wikipedia, and translate them into each target language. The sample is sized to contain the same number of subword tokens as the original pretraining data. We combine the original pretraining data and translated data to create a new set of sentences for continued pretraining, doubling the size of the original. We also finetune the original adapted models using translate-train. We present results in Table \ref{tab:mlm_aug_results}. When finetuning on English and translate-train data, the average performance is highest when using the models adapted on the original data. When finetuning on Spanish, the models adapted on augmented data are best on average. While on average performance increases are not drastic, for some languages the performance increase is notable, and these mixed and/or augmented models may be worth looking into when interested in a particular language.

\begin{table*}[ht]
    \centering
    \small
    \setlength{\tabcolsep}{2.9pt}

    \begin{tabular}{l|ccccccccccc|c}
        \toprule
        & \textbf{FT} & \textbf{aym} & \textbf{bzd} & \textbf{cni} & \textbf{gn} & \textbf{hch} & \textbf{nah} & \textbf{oto} & \textbf{quy} & \textbf{shp} & \textbf{tar} & \textbf{Avg.}\\
        \midrule
        \multicolumn{1}{l}{\textit{Zero-shot}} \\
        \midrule
         XLM-R Large (en) & 89.04 & 40.67 & 41.33 & 43.07 & 42.93 & 39.20 & 45.39 & 42.25 & 42.13 & 48.27 & 40.53 & 42.58 \\
        XLM-R Large (es) & 89.84 & 38.67 & 41.60 & 41.20 & 42.00 & 37.20 & 41.46 & 42.38 & 41.33 & 43.47 & 36.00 & 40.53 \\
        \midrule
        \multicolumn{1}{l}{\textit{Zero-shot w/ adaptation}} \\
        \midrule
        XLM-R Large \texttt{+MLM} (en) & - & 54.80 & 43.87 & \textbf{46.67} & 59.87 & 43.60 & 43.36 & \textbf{44.79} & 64.80 & 43.07 & 41.73 & \textbf{48.66} \\
        XLM-R Large \texttt{+MLM} (es) & - & \textbf{54.93} & 40.40 & 42.93 & 61.07 & \textbf{44.67} & 45.53 & 42.51 & \textbf{68.00} & 43.60 & 40.40 & 48.40 \\
        \midrule
        \multicolumn{1}{l}{\textit{Translate-train}} \\
        \midrule
        XLM-R Large & - & 51.47 & \textbf{50.13} & \underline{33.33} & \textbf{61.20} & 42.00 & \textbf{55.28} & \underline{33.42} & 61.47 & \textbf{49.87} & \textbf{43.87} & 48.20  \\
        \midrule
        \multicolumn{1}{l}{\textit{Translate-test}} \\
        \midrule
        XLM-R Large & - & 38.67 & 40.93 & 35.73 & 50.80 & 38.93 & 39.97 & 32.62 & 47.87 & 39.33 & 35.60 & 40.05 \\

    \bottomrule
    \end{tabular}

    \captionof{table}{Results when using XLM-R Large. Underlined results indicate runs which did not converge on the training data. }
    \label{tab:xlmr_large_results}
\end{table*}
\begin{table*}[ht]
    \centering
    \small
    \setlength{\tabcolsep}{2.9pt}

    \begin{tabular}{l|ccccccccccc|c}
        \toprule
        & \textbf{FT} & \textbf{aym} & \textbf{bzd} & \textbf{cni} & \textbf{gn} & \textbf{hch} & \textbf{nah} & \textbf{oto} & \textbf{quy} & \textbf{shp} & \textbf{tar} & \textbf{Avg.}\\
        \midrule
        \multicolumn{1}{l}{\textit{Zero-shot}} \\
        \midrule
         English & 4.49 & 4.54 & 1.68 & 5.16 & 3.46 & 2.00 & 2.80 & 4.46 & 4.89 & 7.82 & 4.17 & 4.10 \\
        Spanish & 9.07 & 1.42 & 2.22 & 3.91 & 2.75 & 1.38 & 2.48 & 4.06 & 1.82 & 5.07 & 0.27 & 2.54 \\
        \midrule
        \multicolumn{1}{l}{\textit{Zero-shot w/ adaptation}} \\
        \midrule
        \texttt{+MLM} (en) & - & 11.29 & 5.74 & 7.20 & 7.43 & 6.35 & -2.85 & 7.76 & 3.02 & 1.73 & 1.91 & 4.96 \\
        \texttt{+MLM} (es) & - & 11.06 & 0.35 & 4.17 & 8.80 & 6.85 & 1.36 & 1.96 & 5.60 & 3.42 & 1.95 & 4.55 \\

        \midrule
        \textit{Translate-train} & - & 1.47 & -1.29 & -9.12 & 2.31 & -1.20 & -0.05 & -2.59 & 1.56 & -2.13 & 1.83 & -0.92 \\
        \midrule

        \textit{Translate-test} & -     & -1.06 & 0.53 & 1.02 & 4.18 & 0.93 & -1.40 & -2.67 & -3.51 & -0.18 & 0.44 & -0.17 \\

    \bottomrule
    \end{tabular}

    \captionof{table}{Difference in performance between XLM-R Large and Base.}
    \label{tab:xlmr_differences}
\end{table*}
\subsection{XLM-R Large}
\label{sec:large}
In this section we provide results for XLM-R Large. Due to computational restrictions, we slightly modify the experimental setup from the main experiments: we use mixed precision training and a more aggressive early stopping patience of 3 evaluation steps. Additionally, we use a learning rate of 5e-6 for all finetuning experiments, as we found that the original learning rate of 2e-5 failed to converge. However, even when using the modified hyperparameters, we experience some instability during training. The zero-shot model trained on Spanish data did not converge with the original random seed, but successfully trained after changing the seed. For translate-train, the models trained on Asháninka and Otomí failed to converge, regardless of the seed used, and further hyperparameter tuning will be required, which we leave for future work. 

In this experiment, we can see that the results are more varied in comparison to the main results. Translate-train achieves the highest performance for five languages, with the adapted models achieving a combined highest performance for the remaining five. On average, the adapted model finetuned on English labeled data achieved the highest performance, followed closely by the other adapted model, and the translate-train model. This indicates that translate-train may be a viable approach when faced with limited compute, but might also have a restrictive upper limit on performance; in contrast, adaptation may allow for more potential performance gain, especially when larger models and datasets are available. Interestingly, when considering average performances across only the languages for which all models converged (i.e. removing Asháninka and Otomí from the calculation), we find that translate-train offers an average performance of 51.91\%, while adaptation approaches achieve 49.39\% and 49.83\% accuracy on average.

Comparing XLM-R Large to XLM-R Base in Table \ref{tab:xlmr_differences}, we see that for all but one language the Large model outperforms the Base model in all adaptation and zero-shot runs. Notably, the Base model trained on translated data outperforms the Large model, and retains the highest overall performance across all languages and models. 

\clearpage
\renewcommand{\thetable}{D.\arabic{table}}
\setcounter{table}{2}
\begin{table*}
    \centering
    \small
    \setlength{\tabcolsep}{4pt}

    \begin{adjustbox}{width=\textwidth}
    \begin{tabular}{cp{0.9\linewidth}}
        \toprule
        \textbf{Language} & \textbf{Example} \\

\midrule
\multirow{4}{*}{aym}& \textbf{P:}  Mä jan walt ’ awinakax utjkaniti? \\
& \textbf{H:}  Iglesia JI JI ukax XIFlo XICI ukax XIIII ukan mä jach 'a pacha. \\
\cmidrule{2-2}
& \textbf{P:}  Aka qillqatax Crownwn Squareareareare ukax iwayi, ‘ Ñalacio ‘ ‘ ‘ ñoquis ukch ’ añataki. \\
& \textbf{H:}  Plaza de Plaza de palacio palacio palacio äwipat uñt 'ayi. \\
\midrule
\multirow{4}{*}{bzd}& \textbf{P:}  Ye'r ye' alà alà dör ye' alà tã' alàshshshshshshöö ? \\
& \textbf{H:}  Káxkkk e' tã káx batà batà ã káx batà ã . \\
\cmidrule{2-2}
& \textbf{P:}  Káx i'r i' i' ã káx i' ulàshshshshshshshshshshshshshshshshsh . \\
& \textbf{H:}  Kéqéqwöwöwöwöwöwöwö ulà ulà ulà ulà wa . \\
\midrule
\multirow{6}{*}{cni}& \textbf{P:}APAPAPAPAPAPAPAPAPAPAPAPAPAPAPAPAPAPAPAPAPAPAPAPAPAPAPAPAPAPAPAP APAPAPAPAPAPAPAPAPAPAPA))))))))))))))))))))))))))))))))))))))))))))))))))))))))))))))))))))))))))))))) ))))))))))))))))))))))))))))))))))))))))))) O O O O O O O ObibibibibibibiIIIIIIIIIIIIIIIIIII \\
& \textbf{H:}  Ibibibibiti obibiti obibi. Ababababa \\
\cmidrule{2-2}
& \textbf{P:}  b. Akobiro ayiro ayiro ayiro nija Jebabentirori Anampiki. \\
& \textbf{H:}  Itititititititititititi. \\
\midrule
\multirow{4}{*}{gn}& \textbf{P:}  Peteî paseo corto imbarete caminata norte gotyo ha'e pueblo j? Sus , peteî tupão particularmente siglo XIX . \\
& \textbf{H:}  Tupão tavaguasu Jesús omopu'ã siglo XIX . \\
\cmidrule{2-2}
& \textbf{P:}  Péicha Crown Square oime palacio real , kuimba'e preciado tetãme , joy Escocia . \\
& \textbf{H:}  Plaza de la corona cuenta palacio real . \\
\midrule
\multirow{4}{*}{hch}& \textbf{P:}  xewit+ta m+k+ wa+ka xewit+ x+ka xewit+ x+ka mu'at+a. \\
& \textbf{H:}  'aix+ 'aix+ ti'at+ x+t+ x+a mu'at+ x+a. \\
\cmidrule{2-2}
& \textbf{P:}  wa+ka m+k+ 'aix+ pureh+k+t+a de oro. \\
& \textbf{H:}  'ik+ p+h+k+ palacio palacio palacio palacio. \\
\midrule
\multirow{5}{*}{nah}& \textbf{P:}  See tosaasaanil , see tosaasaanil , see tosaasaanil . See tosaasaanil , see tosaasaanil , see tosaasaanil . \\
& \textbf{H:}  Yn ipan ciudad de Jesús la Yglesia de Jesús yn ipan in omoteneuh xihuitl de Jesús . \\
\cmidrule{2-2}
& \textbf{P:}  Auh ynic patlahuac cenpohualmatl ypan in yn oncan tecpan quiyahuac yn oncan tecpan quiyahuac yn tecpan quiyahuac yn oncan tecpan quiyahuac . \\
& \textbf{H:}  In tlapoalli ica tlapoalli ica tecpan palacio . \\
\midrule
\multirow{6}{*}{oto}& \textbf{P:} Ra nge'a mi b'e\b{t}'e\b{m}'i ha ra thuhu ra thuhu ra thuhu ra thuhu ra ñ'o\b{t}'e\b{t}'e\b{t}'a ra thuhu ra thuhu ra thuhu ra thuhu ra thuhu ra thuhu ra thuhu ra hnini . \\
& \textbf{H:}  Nu'u xki tso\b{h}o nuni M'onda . \\
\cmidrule{2-2}
& \textbf{P:}  Ra nge'a ra thuhu ra b'u\b{i} ha ra thuhu ra thuhu ra thuhu ra thuhu ra thuhu ra thuhu ra thuhu ra thuhu ra thuhu ra thuhu ra thuhu ra thuhu ra thuhu ra thuhu , \\
& \textbf{H:}  Ra nge'a ra b'e\b{m}'e\b{m}'e\b{m}'i . \\
\midrule
\multirow{5}{*}{quy}& \textbf{P:}  Asiria nacionpa norte lawninpim, Sus X00 watapa norte lawninpi kaq Sus X00 watakunapi religionniyoq punta apaqkunawan hukllawakurqaku. \\
& \textbf{H:}  Jesuspa tiemponpi iglesia \\
\cmidrule{2-2}
& \textbf{P:}  Crown Squarepa hichpanpim tarikunku palaciopi, chayqa Escocia nacionpa chawpinpim kachkan \\
& \textbf{H:}  Alemania nacionpa Plaza sutiyoq runam qollqepaq apuestaspa palaciopi cuentallikun \\
\midrule
\multirow{8}{*}{shp}& \textbf{P:}  Westiora yoxan yoxanya riki ea, jainxon westiora westiora westiora westiora westioraya iki. \\
& \textbf{H:}  Iririririririririririririririririricancancancancancancan. \\
\cmidrule{2-2}
& \textbf{P:}  Nato yobinbinki jawe ati iki, jainxon min keni raometi iki, jainxon westiora westiora westiora raomeomeai, jainxon min kenkin. \\
& \textbf{H:}Chomomomomomomomomomomomomomomomomomomomomomomomomomomomomomomo momomomomomomomomomomomomomomomomomomomomomomomomomomomomomomomomomomomomomom omomomomomomomomomomomomomomomomomomomomomomomomomomomomomomomomomomomomomomomomomomomomomomomomomomomomomomomomomomomomomin \\
\midrule
\multirow{10}{*}{tar}& \textbf{P:}  ( 2 ) ¿ chí mu ŕe'pá ? ¿ chí mu ŕe'pá ? ¿ atza be'pá ? ¿ chí mu ŕe'pá ? \\
& \textbf{H:}  a'rí ko'rí ko'rí ko'rí ko'rí ko'rí ko'rí ko'rí ko'rí ko'rí ko'rí ko'rí ko'rí ko'rí ko'rí ko'rí ko'rí ko ba'rí ko'rí ko pe pe pe pe pe pe pe pe pe pe pe pe pe pe pe pe pe pe pe a'rí mi mi mi mi mi mi mi mi mi mi mi mi mi mi mi mi'rí ko'rí ko'rí ko ŕe'rí ko'rí ko'rí ko'rí ko'rí ko'rí ko'rí ko'rí ko'rí ko'rí ko mapu mapu mapu mapu ŕe'rí ko'rí ko'rí ko'rí ko'rí ko'rí ko'rí ko ŕe'rí ko ŕe'rí ko ŕe'rí ko'rí ko ŕe'rí ko ŕe'rí ko ŕe'rí ko'rí ko ŕe'rí ko ba'rí \\
\cmidrule{2-2}
& \textbf{P:}  ( 2 ) a ) pe ŕe'pá ŕe'pá ŕe'pá ŕe'pá ŕe'pá ŕe'pá ŕe'pá ŕe'pá ŕe'pá ŕe'pá ŕe'pá ŕe'pá ŕe'pá ŕe'rí ko , pe pe pe ŕe'pá ŕe'pá ŕe'pá ŕe'pá ŕe'pá ŕe'pá ŕe'pá ŕe'pá ŕe'pá ŕe'pá ŕe'pá ŕe'pá ŕe'pá ŕe'pá ŕe'pá ŕe'pá ŕe'pá ŕe'pá ŕe'pá ŕe'pá ŕe'pá ŕe'pá ŕe'pá ŕe'pá ŕe'pá ŕe'pá ŕe'pá bo'pá bo'pá ŕe'pá ŕe'pá ŕe'pá ŕe'pápápá ŕe'pá ŕe'pá ŕe'pá ŕe'pá ŕe'pá ŕe'pá ŕe'pá ŕe'pá ŕe'pá ŕe'pá ŕe'pá ŕe'rí ko ba'pá ŕe'pá ŕe'pá \\
& \textbf{H:}  ( 2 ) ( b ) pe ŕe'chí na'chí na'chí \\

    \bottomrule
    \end{tabular}
    \end{adjustbox}
    \captionof{table}{Two randomly selected translate-train examples.}
    \label{tab:translate_train_sample}
\end{table*}
\end{document}